\documentclass{article} 
\usepackage{iclr2022_conference,times}

\usepackage{amsmath,amsfonts,bm}

\def\eqref#1{equation~\ref{#1}}

\def\1{\bm{1}}

\DeclareMathAlphabet{\mathsfit}{\encodingdefault}{\sfdefault}{m}{sl}
\SetMathAlphabet{\mathsfit}{bold}{\encodingdefault}{\sfdefault}{bx}{n}

\usepackage{hyperref}
\usepackage{algorithm}
\usepackage{algpseudocode}
\usepackage{cite}
\usepackage{amsmath,amssymb,amsfonts}

\usepackage{bbm} 
\usepackage{csquotes}
\usepackage{graphbox}
\usepackage{graphicx}
\usepackage{textcomp}
\usepackage{todonotes}
\usepackage{xcolor}
\usepackage{caption}
\usepackage{subcaption}
\usepackage{wrapfig}
\usepackage{url}
\usepackage{array}
\usepackage{tabularx}

\title{McXai: Local model-agnostic explanation\\ as two games}

\author{Yiran Huang, Nicole Schaal, Michael Hefenbrock, Yexu Zhou, Till Riedel, \\Likun Fang, Michael Beigl \\
Telecooperation Office\\
Karlsruhe Institute of Technology\\
Karlsruhe, Germany \\
}

\iclrfinalcopy 
\begin{document}

\maketitle

\begin{abstract}
To this day, a variety of approaches for providing local interpretability of black-box machine learning models have been introduced. Unfortunately, all of these methods suffer from one or more of the following deficiencies: They are either difficult to understand themselves, they work on a per-feature basis and ignore the dependencies between features and/or they only focus on those features asserting the decision made by the model. To address these points, this work introduces a reinforcement learning-based approach called Monte Carlo tree search for eXplainable Artificial Intelligent (McXai) to explain the decisions of any black-box classification model (classifier). Our method leverages Monte Carlo tree search and models the process of generating explanations as two games. In one game, the reward is maximized by finding feature sets that support the decision of the classifier, while in the second game, finding feature sets leading to alternative decisions maximizes the reward. The result is a human friendly representation as a tree structure, in which each node represents a set of features to be studied with smaller explanations at the top of the tree. Our experiments show, that the features found by our method are more informative with respect to classifications than those found by classical approaches like LIME and SHAP. Furthermore, by also identifying misleading features, our approach is able to guide towards improved robustness of the black-box model in many situations. 
\end{abstract}


\section{Introduction}
With the successful application of machine learning-based classification in a growing number of domains, there is an increasingly high demand for understanding the predictive decisions of machine learning models. One concrete motivation for this is the proliferation of machine learning in the natural sciences, where interpretability is a prerequisite to ensure the scientific value of the results. Another is the use of AI in high-risk situations, as dealt with in the draft for an AI act recently proposed by the European Commission~\citep{pressrelease}.


 Feature importance is the most common explanation for classification~\citep{bhatt2020explainable}. In recent years, many importance-based approaches such as LIME~\citep{ribeiro2016should} or SHAP~\citep{lundberg2017unified} have been proposed. However, most of these approaches study each feature separately, ignoring the relationship between them. In addition to that, features that may lead to misclassification are hardly studied individually. 
 We are therefore searching for a model-agnostic explanation algorithm that can answer the following four questions for a given black-box classifier: (1) The importance of each feature (e.g., pixel of an image) or feature set (e.g., pattern like a logo) for the prediction; (2) The relationship between each feature or feature set according to the black-box model; (3) Based on which feature or feature set is the decision made; (4) Which feature or feature set has a negative influence of the prediction \footnote{It is different from adversarial attack, which would not change the distribution of the input instance. The negative features found by McXai are different from the negative features found by LIME, which presents the features that have a negative correlation with the prediction}?

To address these challenges, we present a novel post-hoc explainability~\citep{tjoa2020survey} approach based on reinforcement learning called Monte Carlo tree search for eXplainable artificial intelligent (McXai). It involves modeling the interpretation process as two games.

In a first "classification game", the agent is tasked with finding features essential to support the correct decision of the classifier,
while in a "misclassification game", the agent seeks to identify features to which the classifier is sensitive, i.e., features whose perturbations may lead to misclassifications.

The agents develop their policies based on a search tree representation, which is constructed using Monte Carlo Tree Search (MCTS)~\citep{chaslot2008monte}. The tree structure represents the dependencies between different feature subsets of the input instance. Hence, through playing these games, the agents generate explanations for reward-maximizing strategies, i.e., sequentially selecting features relevant for the specific decisions.

The contributions of this work can be summarized as follows:
(1) To the best of our knowledge, we are the first to generate interpretations of black-box models through reinforcement learning applied to decision games; (2) Given an input instance, we interpret the prediction of the black-box model in terms of tree structures that humans can naturally grasp; (3) Not only the individual features, we can determine the importance of a random feature set and analyze the dependencies between these features; (4) We can find features or feature sets that are insignificant to the target class, however essential to the other classes and optimize the black box model with these findings.

The rest of this work is organized as follows. Section~\ref{sec: related work} reviews previous works on post-hoc explainability and introduces the MCTS algorithm. A detailed description of the proposed McXai approach is given in Section~\ref{sec: methodology}. We demonstrate the performance of the proposed approach through two experiments in Section~\ref{sec: experiment} and draw conclusions in Section~\ref{sec: conclusion}.

\section{Related Work}
\label{sec: related work}
In this section we give an overview of previous work on technologies for interpreting black-box model and briefly introduce the basic Monte Carlo tree search algorithm as a preliminary.

\subsection{Explainable artificial intelligence}
Many approaches have been proposed to explain the model using a way that is easily perceived by humans.
Permutation Feature Importance~\citep{ruder2016overview} analyzes the prediction change by randomly permuting features in the instance. Class Activation Map (CAM)~\citep{zhou2016learning} decompose signals propagated within its algorithm and processes them with the help of the global average pool to provide an analysis of the prediction. Similarly, Layer-wise Relevance Propagation (LRP)~\citep{bach2015pixel} identifies important pixels by running a backward propagation in the neural network. All these methods show the contribution of pixels to the prediction through a heat map. 

In contrast, CLUE~\citep{antoran2020getting} generates explaination with the help of variational autoencoder. MUSE~\citep{lakkaraju2019faithful} produces an explanation in form of a decision tree. It approximates the black-box model with an interpretable model and optimizes against a number of metrics. The Bayesian Rule Lists~\citep{yang2017scalable} method discretize the feature space into partitions and defines a decision logic within each partition using IF-THEN rules.SHAP~\citep{lundberg2017unified} scores the feature importance with the help of subset. Anchors~\citep{ruder2016overview} creates an explanation based on the perturbation of the features of the input instance. This explanation is presented as IF-THEN rules instead of a surrogate model.

Some other approaches use mathematical structure to reveal the mechanisms of machine learning algorithms. TCAV~\citep{kim2018interpretability} is a technique to interpret the low-level representation of neural network layers. However, instead of single features in the instance, TCAV analyzes the advanced concept (feature set), for example 'striped', with concept activation vector. Another famous method is LIME~\citep{ribeiro2016should}, which is developed by the same researchers that proposed Anchors~\citep{ruder2016overview}. It explains any black-box model with local linear approximation around a given prediction. 

All of the aforementioned methods focus on finding evidence to support the prediction of the black-box model. However, they ignore possible causes of the erroneous prediction. In addition to this, CAM requires the black-box model to have a global average pooling layer and the explanation itself is hard to understand. LIME analyze each feature individually, TCAV analyzes the advanced concept, but requires manual assistance.

\subsection{Monte Carlo tree search}


MCTS~\citep{chaslot2008monte} is a popular heuristic-based reinforcement learning~\citep{minsky1954theory} algorithm that is commonly used to predict moves in board games such as chess and Go. It constructs a tree of the search space to estimate the favorability of actions in a given state. Each edge of the tree thereby represents a move/action and each node represents a state of the game. In order to win a game, the agent performs multiple episodes consisting of four phases: selection, expansion, simulation and back propagation. In the selection process, the agent selects child nodes based on a selection policy until a leaf node is reached. In the expansion process, the agent adds one or multiple child nodes to this leaf node and selects one of them according to an expansion policy. This is followed by the roll-out process, where the game is completed by random moves from the selected node until the game reaches a terminal state (win, lose or draw). 
Based on the end state, a reward $r$ is returned. During back propagation, every node from the selected new child to the root node is updated. The algorithm ends when the pre-defined number of episodes has been reached.

\section{Methodology}
\label{sec: methodology}
\newcommand{\vxc}[1]{\ensuremath{\mathrm{\mathbf{#1}}}}
Given a classification data set $\mathcal{D}$ with $n$ different features and $c$ different classes, a black-box classifier $g$ trained with $\mathcal{D}$ is a system without any internal working knowledge. It takes an instance $x
\in \mathbb{R}^n$ from $\mathcal{D}$ as input and outputs the distribution of the classes as a vector $\vxc{g(x) \in [0, 1]^c}$, where $\vxc{g(x)}[i]$ is the probability that the input instance predicted as class $i$ and $\sum_{i=0}^c \vxc{g(x)}[i] = 1$. The prediction of the black-box model is correct,
if the output with the maximum value (probability), i.e. $\underset{i\in\left\{0,\cdots,c\right\}}{\operatorname{argmax}}\; \vxc{g(x)}[i]$, equals the ground-truth class $y \in \mathbb{N}$.

McXai determines the importance of a feature by observing the change of the prediction probability after removing the target feature from the input instance $x$. To keep the shape of the input dimension~\footnote{Most black box models have a fix input dimension (number of features).}, instead of removing the feature, we mask the value of the target feature as a given constant $\tau$, where $\tau$ should not give away any information about the input instance $x$. In the case of image data, it is set to zero or the average value of all the features (pixels) in the images. The specific analysis method is further described in the next subsection.

\subsection{Generating explanations through games}
Given a black-box model $g$, McXai models the interpretation as the classification and misclasification games, where agents 
are given the possibility to mask a feature, e.g., pixel region in an image in each step of the game by changing its value as a given constant $\tau$. Formally, let $x \in \mathbb{R}^{n}$, the action $a \in \{0,1\}^{n}$, with $\sum_{i=0}^n a_i = 1$. When an action $a$ is chosen by an agent, it is applied to $x$ element-wise, which results in the new state $x' = x \odot \Bar{a} + \tau \cdot a$, where $\Bar{a}$ is the logical not of the action $a$. Hence, $a$ masks the value of the target feature as $\tau$. For example, given $x =\left\{1,2,3\right\}$, $a=\left\{0, 1, 0\right\}$ and $\tau=10$, new state
$x'=\left\{1,10,3\right\}$.

\textbf{The classification game} starts, when the given data instance $x$ correctly classified, i.e., $\underset{i\in\left\{0,\cdots,c\right\}}{\operatorname{argmax}}\; \vxc{g(x)}[i]$ = y. In each step, the agent successively masks value of features, i.e. dimensions, of $x$ as $\tau$ through its actions $a$, trying to achieve a misclassification. When a misclassification is achieved, i.e., $\underset{i\in\left\{0,\cdots,c\right\}}{\operatorname{argmax}}\; \vxc{g(x)}[i] \neq y$, the game is terminated. The reward is given based on the number of steps required to terminate the game. This can be understood as trying to find the shortest and most concise description in terms of feature relevance.

\textbf{The misclassification game}, contrary to the classification game, is played for misclassified instances, and it can either be played directly for wrongly labeled instances, or, as a continuation of the classification game (from the final state). In the misclassification game, the goal of the agent is to find the least number of actions (maskings) required to achieve a correct classification. Hence, the game encourages masking features that lead to misclassification. As for classification game, the reward is designed to encourage a fast termination of the game.  
\begin{figure}[t!]
\centering
\newcommand{\wdt}{0.22\textwidth}
\begin{subfigure}[tb]{0.59\textwidth}
\centerline{
\includegraphics[width=1\textwidth]{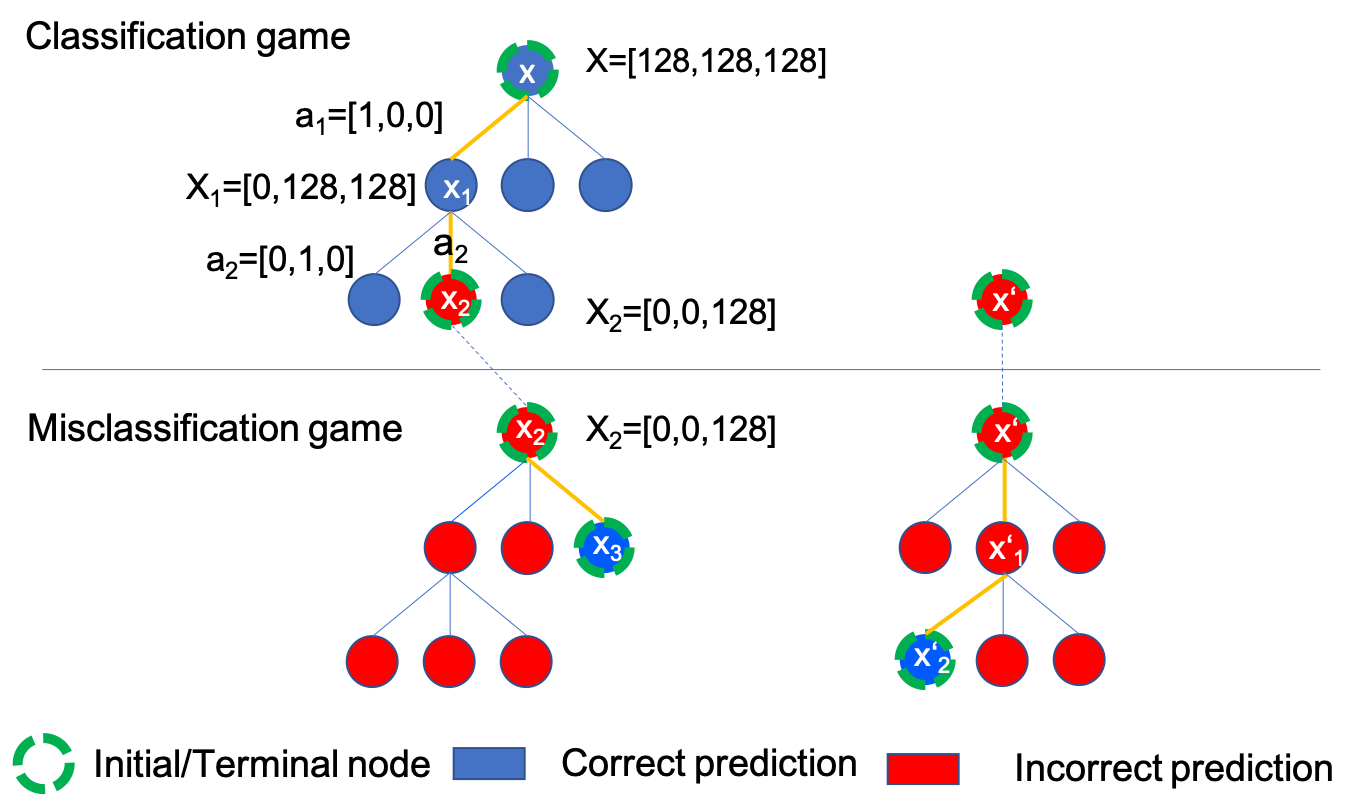}
}
\caption{}
\label{fig: tree representation}
\end{subfigure}
\begin{subfigure}[tb]{0.39\textwidth}
\centering
\includegraphics[width=0.8\textwidth]{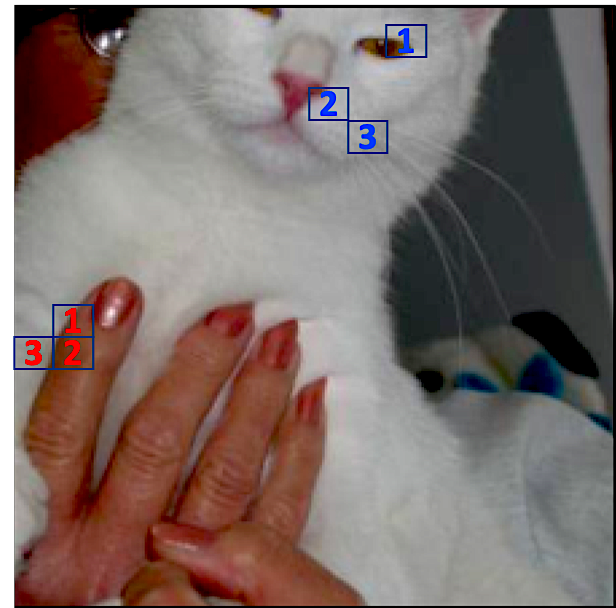}
\caption{}
\label{fig: example_cat}
\end{subfigure}
\caption{(a) Two examples about classification game and misclassification game. $x$ and $x'$ are two instances from the data set that used to train the black-box model, where $x$ is predicted correctly and $x'$ is predicted incorrectly. $\tau$ is set to zero. (b) Example of top three important features for the classification (blue) and misclassification (red) game.}
\end{figure}
Figure~\ref{fig: tree representation} shows two examples of the games. The input instance $x$ contains three features. At the beginning of the classification game, the prediction of the black-box model is correct. Action $a_1$ is applied to $x$, masks the first feature in $x$ and produces a new state $x_1$. Similarly, the state $x_2$ is generated by masking the second feature of $x_1$ with the action $a_2$. The prediction of the black-box model for the $x_2$ is incorrect, thus the classification game ends. The misclassification game starts at the end node of the classification game. At each decision step, an action is selected from the action space and applied to the current state to generate a new state. This operation is repeated until the black-box model's prediction of the new state is correct and the game is declared as over. For the instance $x'$, since the prediction of the black-box model is wrong, we can analyze the instance directly with the misclassification game.

In the following, we first introduce the Monte Carlo tree representation of the explanation result for a given data instance $x$ and the information hidden in the tree. Then we present the selection and expansion policies used to construct the tree. After that, we summarise the McXai algorithms in the last subsection. 

\subsection{Tree representation}
The explanation of a given instance $x \in \mathcal{D}$, is presented as a Monte Carlo tree in McXai. 
Each edge in the tree represents an action $a\in\left\{0,1\right\}^n$ (bit mask), that corresponds to an individual feature. It contains two attributes: How many times the edge has been explored (number of visits); The expected reward for taking the action in the parent node (win rate).
The nodes in the tree represent the states of the game and are divided into three categories: (1) \emph{start (root) node} $x_0$ represents the initial state of the game; (2) \emph{derived nodes} $x_i$ with $i>0$ are the masked instance with only one parent node $x_j$ with $i>j\geq 0$. The edge connecting these two nodes corresponds to the action $a$ applied to the parent node, i.e., $x_{i} = x_j \odot \Bar{a} +\tau \cdot a$; (3) \emph{terminal node} $x_t$ represents the end state of the game.
Each path in the tree corresponds to a feature set, and their importance are presented by the win rate of the last edge in the path. A complete path is a path that connects root node and terminal node. 

From the generated tree structure of the classification game, the following information can be extracted:
\begin{itemize}
    \item \textbf{The importance of each individual feature or feature set to the correct prediction} Here, each feature is considered to be independent of the other features. This information is shown in terms of the win rate of initial state's edges. 
    \item \textbf{The importance of any feature set for the correct prediction} The importance of one feature may depend on the other features in the same set. As with the phenomenon of multicollinearity, the contribution of two features to the prediction is the same. If the two features are considered separately, they have the same importance. However, if they are considered together, one of them is sufficient to make a prediction. Their importance is expressed in terms of the win rate of the corresponding path in the tree.
    \item \textbf{The relationship between different features} Since each path in the tree corresponds to a feature set, the relationship between features is presented as a change in the win rate of the edge in the corresponding path.
    \item \textbf{The most important set of features that proves the correct prediction} The most important feature set is reflected by the complete path with the largest win rate.
\end{itemize}
Similar information can be extracted from the tree of the misclassification game. However, the found features are not the key to supporting the prediction, but for rejecting it.

Figure~\ref{fig: example_cat} shows an example of the top three features found by McXai for both games~\footnote{Image comes from the dog-cats open source data set. The black-box model is an 18-layer ResNet~\citep{he2016deep}. For the sake of clarity, we have shown only the three most important features.}. The black-box model predicts that the instance is a cat, based primarily on the features marked in blue: the eye and whisker of the cat. However, we see that, although the whisker serve as an important evaluation criterion, the whiskers on both sides are not equally evaluated, as multiple features are considered together. This differentiation is black-box determined, and many explanation methods fail to detect this. We demonstrate this in 
section~\ref{sec: experiment}. The black-box model predicts the original instance of a cat with a 93\%, which dropped to 71\% after removing the blue features. This reflects the important role played by these features in correct prediction. Human skin (the features marked as red) is considered to have a negative effect on the correct prediction. The removal of these attributes restore the probability of the prediction to 80\%. 
It is also worth noting that the excessive focus on a small amount of useful features e.g., eye of the cat, can be the reason that limits the performance of the black-box model. If we can reduce the attention on these 'important' features during the training process and thus enhance the model's focus on other patterns, the performance of the black-box may be improved. This hypothesis will be explored in Experiment~\ref{sec: experiment}.

\subsection{Exploration policies}
McXai applies MCTS to construct the tree representation. Two policies are defined in the MCTS algorithm, namely selection policy $\pi_s$ and expansion policy $\pi_e$, which, as their names suggest, are applied to the selection and expansion phases of the MCTS algorithm, respectively. 

Similar to general MCTS algorithm, we apply the Upper Confidence Bound for Tree (UCT) as selection policy~\footnote{The feasibility of this approach is theoretical demonstrated in \citet{coquelin2007bandit}.}:
\begin{equation}
    \pi_s(x) = \underset{a \in \mathcal{A}}{\operatorname{argmax}} \quad \left\{\mu_{x,a} + \lambda \cdot \sqrt{\frac{\log n(x)}{n(x,a)}}\right\}
    \label{eq:pi_s}
\end{equation}
where $\mathcal{A}$ is the action space, $\mu_{x,a}$ is the win rate of the edge: taking action $a$ at parent state $x$. Furthermore, $n(x) = \sum_{a\in A}n(x,a)$ denotes the number of visits to state $x$ and $n(x,a)$ signifies the number of visits to the edge.
The parameter $\lambda$ is used to adjust the trade-off between the number of visits and the win rate. It is set to $\sqrt{2}$ to ensure the asymptotic optimum for the MCTS algorithm, since the reward range of both games is in [0, 1] \citep{kocsis2006bandit}.

Unlike the selection policy $\pi_s(x)$, which directly applies the explored information to guide decisions, an exploration policy first approximates the win rate of unexplored states by training a surrogate model $Q$ with explored information. The action $a$ with the maximal predicted win rate is then selected. The surrogate model $Q$ takes a state $x$ and an action $a$ as input and predicts the win rate of taking action $a$ at the state $x$. The expansion strategy can therefore be summarised as:

\begin{equation}
    \pi_e(x) = \underset{a \in \mathcal{A}}{\operatorname{argmax}}\quad Q(x, a)
    \label{eq:pi_e}
\end{equation}

The choice of the surrogate model depends on the type of the data set. For image data we apply a network applied by the policy network of AlphaGo~\citep{longo2020explainable}.


The surrogate model is trained with the 'win rate' attribute stored in each edge of the tree, which is updated in each episode after the back propagation as proposed by~\citet{chaslot2008monte}.

\subsection{Tree construction for games} 
For classification game, the black-box model predicts the initial state correctly. The importance of a feature can be observed by comparing the change in the prediction probability of the target class before and after the corresponding feature is masked in the input instance. If the feature is important to the prediction, removing it should significantly reduce the prediction probability. Conversely, removing unimportant features should have little impact. Naturally, if the chosen features are important, fewer actions are required to terminate the game.
A similar situation also exists for misclassification game. However, if a chosen feature is important (for the misclassification), removing it will, conversely, increase the prediction probability of the target class.

The algorithm of the tree construction follows the general MCTS scheme~\citep{chaslot2008monte}, where the termination and reward function are modified. Given a black-box model $g$, an initial node $x_0$ with target $y$ and a terminal node $x_t$, the reward function $r(\cdot)$ is defined as:
\begin{equation}
    r(x_t) = \left[\left(1-\eta\right) \cdot \left(1-\frac{l(x_t)}{L}\right) +\eta \cdot  q\right] \cdot \mathbbm{1}_{\{l(x_t) \le L\}}
    \label{eq:reward1}
\end{equation}
with 
\begin{equation*}
\begin{split}
    &p = \underset{i\in\left\{0, \cdots,c\right\}}{\operatorname{argmax}} \; g(x_0)[i]\\
    &q = \left(2 \cdot \mathbbm{1}_{\{p=y\}}-1\right) \cdot (g(x_0)[y] - g(x_t)[y])
\end{split}
\end{equation*}
$\mathbbm{1}$ is the indicator function, which is used to identify the type of the game. $p$ is the predicted class of the black-box model $g$ given $x_0$. 
If it is the classification game, $p$ should be the same as the target $y$, otherwise, it is the misclassification game. 
In the case of misclassification game, the probability of the initial node (of the target category) is smaller than the probability of the target node. $(g(x_0)[y] - g(x_t)[y])$ leads to a negative value. With the help of indicator function, we can turn it into a positive value. 
$l(\cdot)$ returns the depth of the given node (the number of actions taken to reach the given state.) and $L$ is the maximal depth of the tree, which is used to limit the size of the tree. Also, the maximum depth $L$ increases the value represented by the number of actions. The smaller the value $L$, the greater the reward for the same depth $l$. Thereby, small step differences will appear more pronounced. $q$ denotes the difference between the prediction probability of the target class before and after removing the selected features.
$\eta \in [0,1]$ is a parameter, that is used to weight the path length and the probability change. We can see that when the depth of the terminal node is less than $L$, the reward is a weighted sum of the depth and the probability change. When the depth is larger then $L$, which means that the number of actions required to end the game exceeds a threshold, we set the reward to zero to reduce the likelihood of these actions (actions taken to reach the terminal node) being selected again.

MCTS constructs the tree iteratively. Each iteration involves following four phases:
\begin{itemize}
    \item \textbf{Selection} MCTS traverses the tree from the root node according to \eqref{eq:pi_s}, until reaching a leaf node.
    \item \textbf{Expansion} A new child node is selected according to \eqref{eq:pi_e} and added to the tree.
    \item \textbf{Roll-out} A new action is selected randomly and applied to the current node until reaching a terminal node or the maximal depth of the tree. Then associated reward is computed according to \eqref{eq:reward1}.
    \item \textbf{Back propagation} The reward is back-propagated along the current path, incrementing 'number of visits' and adding reward to 'win rate' of all visited edges.
\end{itemize}

Because of the tree structure, the time complexity of our algorithm is in $O(|\mathcal{A}| I \log |\mathcal{A}|)$ where $I$ is the number of performed episodes. The space complexity is in $O(|\mathcal{A}| \log |\mathcal{A}|)$.


\section{Experiments}
\label{sec: experiment}

In this section, we experimentally demonstrate the capabilities of our proposed approach in interpreting a black-box model in two ways: comparing the importance of positive features found by classification game with classical post-hoc explainability methods; measuring the improvement of the black box model after retraining with the features found in misclassification game.


\subsection{Classification game: Comparing features with positive impact}
We hypothesize that considering the individual dependencies between features in an explanation increases the quality of this explanation. In the first experiment, we want to validate this assumption. To this end, we designed a task on the open source MNIST dataset and all the sklearn real world classification dataset: covertype, kddcup, newsgroup, face\footnote{RCV1 dataset is no included because it is multi label classification task.}. We compared the performance of LIME~\citep{ribeiro2016should}, SHAP~\citep{lundberg2017unified}\footnote{We select two most famous general post-hoc methods, since the data sets used here are not limited to image and some black-box models do not provide gradient information.} and the proposed McXai model on this task. To demonstrate the generality of the method, different kinds of models are trained for these data set. 
For the McXai, the hyperparameter $\tau$ is set to zero and $\eta$ is set to 0.5, so that the probability and the path length are treated equally. Besides, the maximal depth $L$ of the constructed tree is set to 10 to limit the size of the built tree, except for newsgroup dataset, which maximal depth is set to 30 since it has a significant larger number of features. Totally 50 instances are randomly selected from each data set and used as input of the task. 

The task design is inspired by the experiment in \citep{lundberg2017unified}. Taking MNIST data set for example, the task can be described as follows: given an instance from the MNIST data set with target class $y=7$, features of the instance are continuously removed from the instance according to the proposal of the XAI model, so that the prediction of the instance (of the black-box model) is converted to any other class. The post-hoc  explanation approach contributes to the task by analyzing the input instance and extracting a list of features ranked according to their importance.
To compare the importance of the found features by each algorithm, we measure the number of steps (NoS) that are required to change the predicted class according to the feature list of each method. The idea is that the fewer steps are needed, the more important are the features found by the corresponding explainability approach. 

The result of the experiment is summarized in Table~\ref{table:exp2}. On all data sets except newsgroup, the proposed McXai algorithm achieves optimal results. SHAP has the best results on the news data set. However, for each instance in this data set, SHAP takes on average five minutes to analyze, three minutes for McXai. LIME is the fastest, yet with the worst result. The runtime of the McXai is influenced by following three factors of the data set: 1) Number of features, the more features the data set contains, the longer the run time. It takes on average 30 seconds to analyse an instance of 'covertype' data set,  while three minutes for instance in newsgroup. 2) Complexity of the relationship between features and prediction. It is measured by the number of modified features needed to change the model prediction. The lower the number, the lower the corresponding complexity and therefore the shorter the running time. In contrast to the 'newsgroup' data set, McXai takes an average of five minutes to analyze one instance in the dog-cat data set, which contains 16384 features, since the data set is more complicated. 
3) Complexity of the black-box model. The more complex the black box model, the more time takes McXai.

\begin{table}[htbp]
\caption{Comparing average number of steps (NoS) needed to take to change the prediction of black-box model according to the suggestion of the LIME, SHAP and McXai methods.}
\begin{center}
\begin{tabular}{|c|c|c|c|c|c|}
\hline
\textbf{Dataset} & \textbf{Type} & \textbf{No. features}& \textbf{NoS-Lime} & \textbf{NoS-SHAP} &\textbf{NoS-McXai}\\
\hline
\hline 
\textbf{MNIST} & image & 784 & 7.23 $\pm$ 5.65 &6.23 $\pm$ 5.34 & \textbf{4.82 $\pm$ 2.65}\\
\hline
\textbf{covertype} & relational & 54 &10.32 $\pm$ 3.36 & 1.72 $\pm$ 0.82 & \textbf{1.59 $\pm$ 0.88}\\
\hline
\textbf{kddcup} & relational & 41 & \textbf{1.0 $\pm$ 0} & 2.2 $\pm$ 2.0 & \textbf{1.0 $\pm$ 0}\\
\hline
\textbf{newsgroup}& text & 15698 & 46.26 $\pm$ 36.8 & \textbf{6.69 $\pm$ 7.78} &  8.2 $\pm$ 3.72\\
\hline
\textbf{face} & image &4096 & 24.94 $\pm$ 7.76 & 17.62 $\pm$ 6.67 &\textbf{5.62 $\pm$ 0.82}\\
\hline
\end{tabular}
\label{table:exp1}
\end{center}
\end{table}

\begin{figure}[t!]
\centering
\newcommand{\wdt}{0.22\textwidth}
\begin{subfigure}[tb]{0.49\textwidth}
		\begin{tabularx}{\textwidth}{XXXX}
			LIME 
			&
			\includegraphics[align=c,width=\wdt]{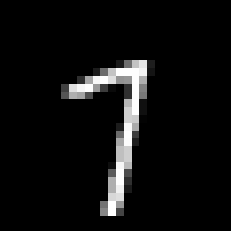} 
			&
			\includegraphics[align=c,width=\wdt]{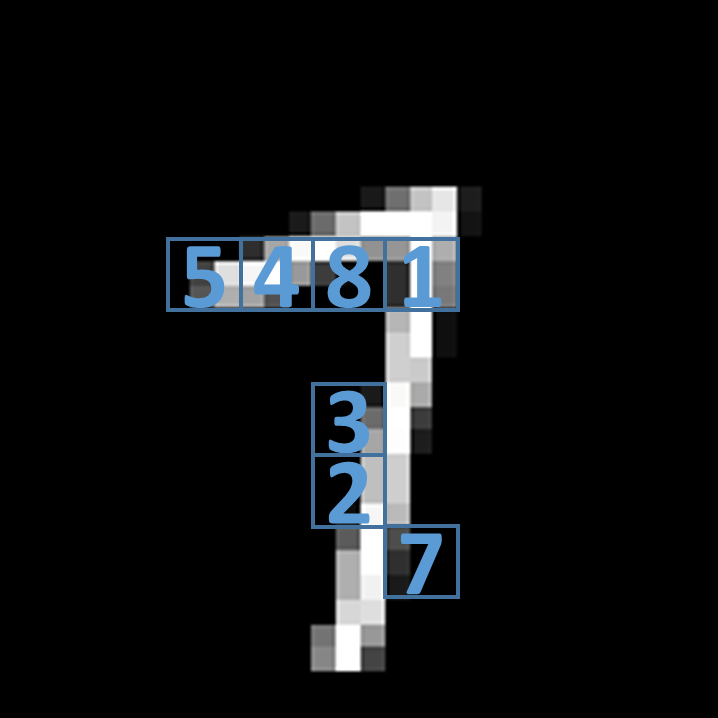} 
			&
			\includegraphics[align=c,width=\wdt]{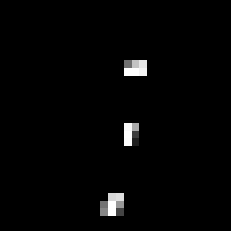} \\[7.5mm]
			SHAP
			&
			\includegraphics[align=c,width=\wdt]{images/original} 
			&
			\includegraphics[align=c,width=\wdt]{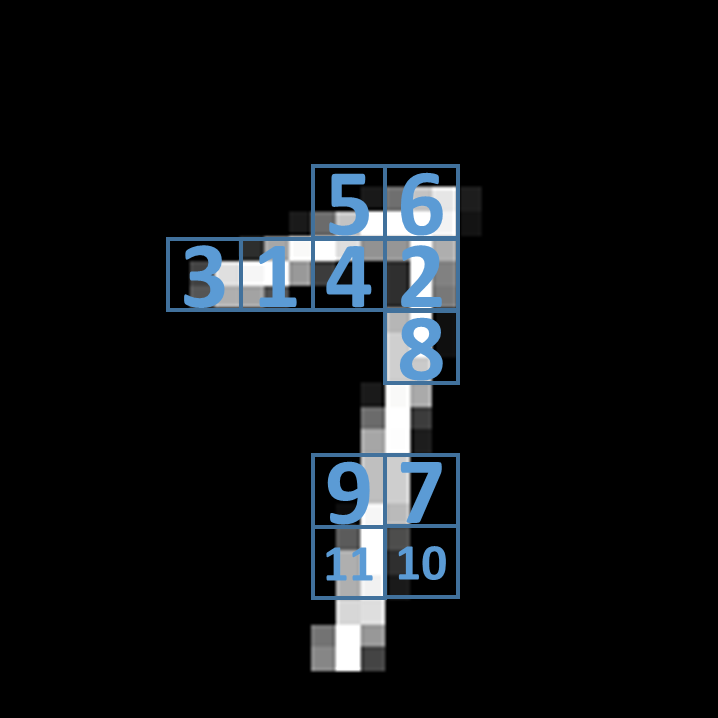} 
			&
			\includegraphics[align=c,width=\wdt]{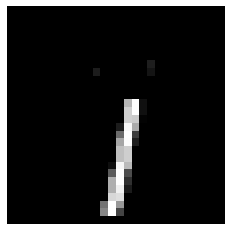} \\[7.5mm]
			McXai
			&
			\includegraphics[align=c,width=\wdt]{images/original} 
			&
			\includegraphics[align=c,width=\wdt]{images/lime_scores} 
			&
			\includegraphics[align=c,width=\wdt]{images/lime_p1} \\[7.5mm]
			& \hspace{5mm} (A) & \hspace{5mm} (B) & \hspace{5mm} (C)
		\end{tabularx}
	\caption{}
	\label{exp1_masking}
\end{subfigure}\hfill
\begin{subfigure}[tb]{0.49\textwidth}

	\centering
	\includegraphics[width=0.8\textwidth]{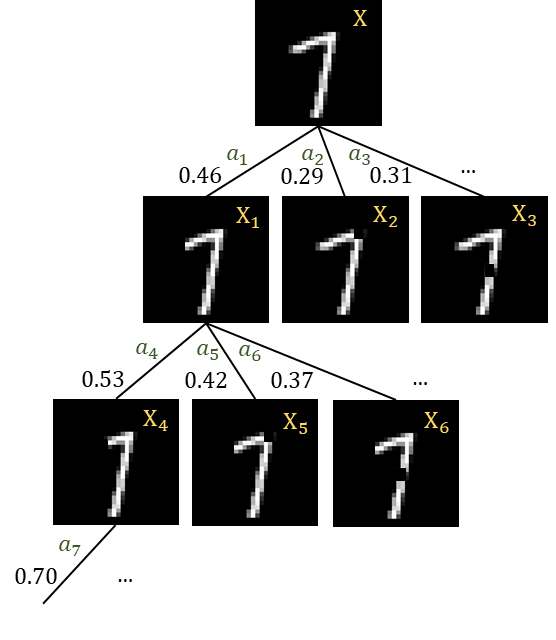}
	\caption{}
	\label{mcxai_tree}

\end{subfigure}\hfill
\caption{(a) Ranks and masked image for each method. (A) Shows the input instance. (B) Shows the explanation created by each algorithm. The features colored in blue have a positive feature importance according to each method. (C) Shows the masked image which is no longer predicted as 7. (b) Shows the tree created by McXai from the example of (a). The state $x$ is the input instance. The state $x_1$ is created by applying the action $a_1$ on $x$. The state $x_4$ is created by applying the action $a_4$ on $x_1$ and so on. The win rate of each action in the considered state is written beside the corresponding edge. The left path is the best path containing the actions with the highest win rate of the considered states.} 
\end{figure}

Figure~\ref{exp1_masking} shows a MNIST example of the features found by the XAI models. In this example, we can see that the feature found by LIME are scattered throughout the number. However, this does not mean that the black-box model makes predictions based on the skeleton of the number, as illustrated by the properties found by SHAP and McXai. This suggests that LIME ignored the competiting (e.g., Feature b is less important when feature a is present.) and conditional (e.g., Feature b is important only when feature a is present.) relationships between features, thus misjudging the importance of some features. Another interesting point is that SHAP and McXai found almost the same features, except for the difference in the ranking. It reflects another attribute of McXai, which ranks the importance of the already explored attributes during the operation of the algorithm (reflected by the selection policy) and prioritizes the exploration of the attributes with higher importance. The corresponding tree is partly shown in Figure~\ref{mcxai_tree}. From this experiment, we can conclude that the positive effect of the features found by McXai is more significant than the other two methods.

\subsection{Misclassification game: Testing the improvement of black box model through retaining}

The misclassification game identifies features that are insignificant to the target class, but sensitive to other classes. In practice it manifests itself as the reduction in the probability of an instance being correctly predicted by adding corresponding features to it. We hypothesize that, if the misclassification game can find these features with negative impact, by counteracting the effect of these features, the performance of black-box model will be improved. To this end, we designed the following experiment on the open source dog-cats data set from Kaggle:

Firstly, we split the dog-cats data set into training set $\mathcal{D}_{train}$ and testing set $\mathcal{D}_{test}$, train a black-box model with $\mathcal{D}_{train}$ and record the accuracy of the black-box model.
Secondly, we analyze each instance in the training set with the proposed approach, where parameters $\eta$ is set 0.5, constant $L$ is 20 and $\tau$ is zero.
Thirdly, for each instance in the training set, we remove the features found by misclassification game to form a new data set $\mathcal{D}_{mis}$.
Fourthly, for each instance in the training set, we remove the features found by both games to form a new data set $\mathcal{D}_{both}$.
Then, we retrain the black-box model with $\mathcal{D}_{train}$ and $\mathcal{D}_{mis}$ and record the accuracy of the black-box model and retrain the black-box model with $\mathcal{D}_{train}$ and $\mathcal{D}_{both}$ and record the accuracy of the black-box model.
Finally, we compare the performance of the three above trained models on $\mathcal{D}_{test}$.

We run the above experiment on the following five torchvision pretrained models: (1) MnasNet~\citep{tan2019mnasnet} with depth multiplier of 0.5 (mnasnet0\_5); (2) MnasNet with depth multiplier of 1.0 (mnasnet1\_0); (3) DenseNet121~\citep{huang2017densely} (4) WideResNet~\citep{zagoruyko2016wide}; (5) GoogleNet~\citep{he2016deep}. Since all these models are pretrained with the ImageNet data set and therefore converge quickly in the experiment\footnote{All models converged after two to three episodes, except for mnasnet1\_0 which took seven episodes in average. The size of the data therefore has little impact on the results.}.

For each model we train with 20 epochs and repeat this 5 times to record the mean and standard of the performance. The dog-cats data set consists of 2500 training images and 500 test images, where the images of cats and dogs are equally represented in the training and the test set. For this reason, we use the accuracy to measure the performance of the models.

The results of the experiment are summarised in Table~\ref{table:exp2}. Overall, removing the features found by misclassification game leads to an improvement in accuracy or stability of the model. This shows that misclassification game can indeed find factors that lead to incorrect predictions and counteracting their effect can improve the performance of the black-box model. In addition, we can derive other interesting information from the results of this experiment. The accuracy improvement caused by removing the negative features is different from model to model. When the original accuracy of the model is low, this improvement is more pronounced. In the case of a model like GoogleNet, which original accuracy is high, the improvement in accuracy is almost untraceable (less than 1\%), however the improvement in stability is significant.  Also, as shown in the result, removing the features found by both classification game and misclassification game wins twice in the experiment, which proves the hypothesis we discussed in Section~\ref{sec: methodology}. 
As classification game finds the features that are considered most important for the correct prediction of the black-box model, removing these features from the training set exacerbates the influence of other features in the input, thereby increasing the generality of the black-box model. 

\begin{table}[htbp]
\caption{Comparing performance of black-box model: mnasnet0\_5, mnasnet1\_0, DenseNet121, WideResNet and GoogleNet in these three different situations: (1) trained with training set $\mathcal{D}_{train}$ (base\_score) (2) trained with training set $\mathcal{D}_{train}$ and $\mathcal{D}_{both}$ (score\_both) (3) trained with training set $\mathcal{D}_{train}$ and $\mathcal{D}_{mis}$ (score\_mis)}
\begin{center}
\begin{tabular}{|c|c|c|c|c|}
\hline
 & \textbf{base\_score (\%)} & \textbf{score\_both (\%)} & \textbf{score\_mis (\%)}\\
\hline
\hline 
\textbf{mnasnet0\_5} & 87.78 $\pm$ 3.31 & 87.98 $\pm$ 1.71 & \textbf{89.16 $\pm$ 3.25}\\
\hline
\textbf{mnasnet1\_0} & 88.65 $\pm$ 3.17 & \textbf{92.83 $\pm$ 1.15} & 91.03 $\pm$ 3.06\\
\hline
\textbf{DenseNet121} & 91.23 $\pm$ 1.46 & \textbf{93.7 $\pm$ 2.63} & 92.98 $\pm$ 2.51\\
\hline
\textbf{WideResNet} & 83.35 $\pm$ 2.25 & 79.49 $\pm$ 4.42 & \textbf{87.58 $\pm$ 4.66}\\
\hline
\textbf{GoogleNet} & 93.97 $\pm$ 3.96 & 94.52 $\pm$ 1.33 & \textbf{94.58 $\pm$ 1.33}\\
\hline
\end{tabular}
\label{table:exp2}
\end{center}
\end{table}

\section{Conclusion}
\label{sec: conclusion}
In this paper, we propose a novel approach called McXai to improve the reliability of black-box models through explaining the principle behind the decision. This method can be used to analyze the classification decision of a black-box model by considering a single feature or a feature set in an input instance in a way that can be easily understood by humans and to find the factors that are positive and negative for the model prediction.

The cornerstone of this approach is to formalize the XAI problem in a reasonable way as two games and to make each game focus on finding specific properties. This simulation allows us to deal with the XAI problem in a way as classic game and learn an agent that handles this problem. Inspired by AlphaGo~\citep{longo2020explainable}, MCTS algorithm is adapted and applied to our task, which enables to present the analysis of the prediction of an input instance in a tree structure while learning the agent. This greatly increases the interpretability.  

In our experiments we compare the found positive features of different XAI approaches and test the found negative features of the proposed approach in different black-box models. Through these experiments we demonstrate the ability of the proposed approach in explaining the prediction of black-box models. Moreover, we found that with the identified features, we can further improve the performance of black-box models.


\bibliography{iclr2022_conference}
\bibliographystyle{iclr2022_conference}

\appendix
\section{Appendix}

\end{document}